\documentclass[10pt,twocolumn,letterpaper]{article}

\usepackage{cvpr}
\usepackage{times}
\usepackage{epsfig}
\usepackage{graphicx}
\usepackage{amsmath}
\usepackage{amssymb}

\usepackage{hhline}
\usepackage[table]{xcolor}
\usepackage{colortbl} 
\usepackage{diagbox}
\usepackage{booktabs}
\usepackage{multirow}
\usepackage{makecell}
\usepackage[table]{xcolor}
\usepackage{graphicx}
\usepackage{subfigure}
\usepackage{multirow}
\usepackage{amssymb}
\usepackage{amsmath}
\usepackage{booktabs}
\usepackage[ruled]{algorithm2e}
\usepackage{color}
\usepackage{ragged2e}
\usepackage{cite}
\usepackage{colortbl}
\usepackage{lineno}
\usepackage{diagbox}
\usepackage{caption}
\usepackage{xspace}
\usepackage{enumitem}
\usepackage{array} 

\usepackage[pagebackref=true,breaklinks=true,letterpaper=true,colorlinks,bookmarks=false]{hyperref}

\cvprfinalcopy 

\def\cvprPaperID{****} 

\ifcvprfinal\fi
\begin{document}

\newcommand{\moduleOneBig}{Dual-perspective Cropping Module\xspace}
\newcommand{\moduleOneBigShort}{DCM\xspace}
\newcommand{\moduleTwoBig}{Dual-perspective Enhancement Module\xspace}
\newcommand{\moduleTwoBigShort}{DEM\xspace}
\newcommand{\modelname}{INF-LLaVA\xspace}

\title{INF-LLaVA: Dual-perspective Perception for High-Resolution Multimodal Large Language Model}

\makeatletter
\def\thanks#1{\protected@xdef\@thanks{\@thanks
        \protect\footnotetext{#1}}}
\makeatother
\author{
    \text{Yiwei Ma \quad Zhibin Wang \quad Xiaoshuai Sun \quad Weihuang Lin \quad Qiang Zhou \quad Jiayi Ji \quad Rongrong Ji} \\
    \texttt {yiweima@stu.xmu.edu.cn}
    \texttt {xssun@xmu.edu.cn}
}


\maketitle

\begin{abstract}
With advancements in data availability and computing resources, Multimodal Large Language Models (MLLMs) have showcased capabilities across various fields.
However, the quadratic complexity of the vision encoder in MLLMs constrains the resolution of input images.
Most current approaches mitigate this issue by cropping high-resolution images into smaller sub-images, which are then processed independently by the vision encoder.
Despite capturing sufficient local details, these sub-images lack global context and fail to interact with one another.
To address this limitation, we propose a novel MLLM, \modelname, designed for effective high-resolution image perception.
\modelname incorporates two innovative components.
First, we introduce a \moduleOneBig (\moduleOneBigShort), which ensures that each sub-image contains continuous details from a local perspective and comprehensive information from a global perspective.
Second, we introduce \moduleTwoBig (\moduleTwoBigShort) to enable the mutual enhancement of global and local features, allowing \modelname to effectively process high-resolution images by simultaneously capturing detailed local information and comprehensive global context.
Extensive ablation studies validate the effectiveness of these components, and experiments on a diverse set of benchmarks demonstrate that \modelname outperforms existing MLLMs.
Code and pretrained model are available at \url{https://github.com/WeihuangLin/INF-LLaVA}.
\end{abstract}


{\section{Introduction}\label{sec:intro}}

The field of multimodal large language models (MLLMs)~\cite{liu2024visual,sun2024generative,dai2024instructblip} has achieved substantial breakthroughs, driven by monumental advancements in computer vision~\cite{he2016deep,han2022survey,dosovitskiy2020image} and natural language processing~\cite{zhao2023survey,achiam2023gpt,touvron2023llama}.
These MLLMs have demonstrated exceptional efficacy across an array of complex tasks, including image captioning~\cite{cornia2020meshed,luo2021dual}, visual question answering~\cite{antol2015vqa,xu2016ask}, and visual dialogue~\cite{das2017visual,massiceti2018flipdial}. This substantial progress not only highlights the transformative potential of MLLMs but also significantly extends the boundaries of understanding, reasoning, and interactive capabilities integral to the development of general artificial intelligence (AGI).

\begin{figure}[]
\centering
\includegraphics[width=1.0\columnwidth]{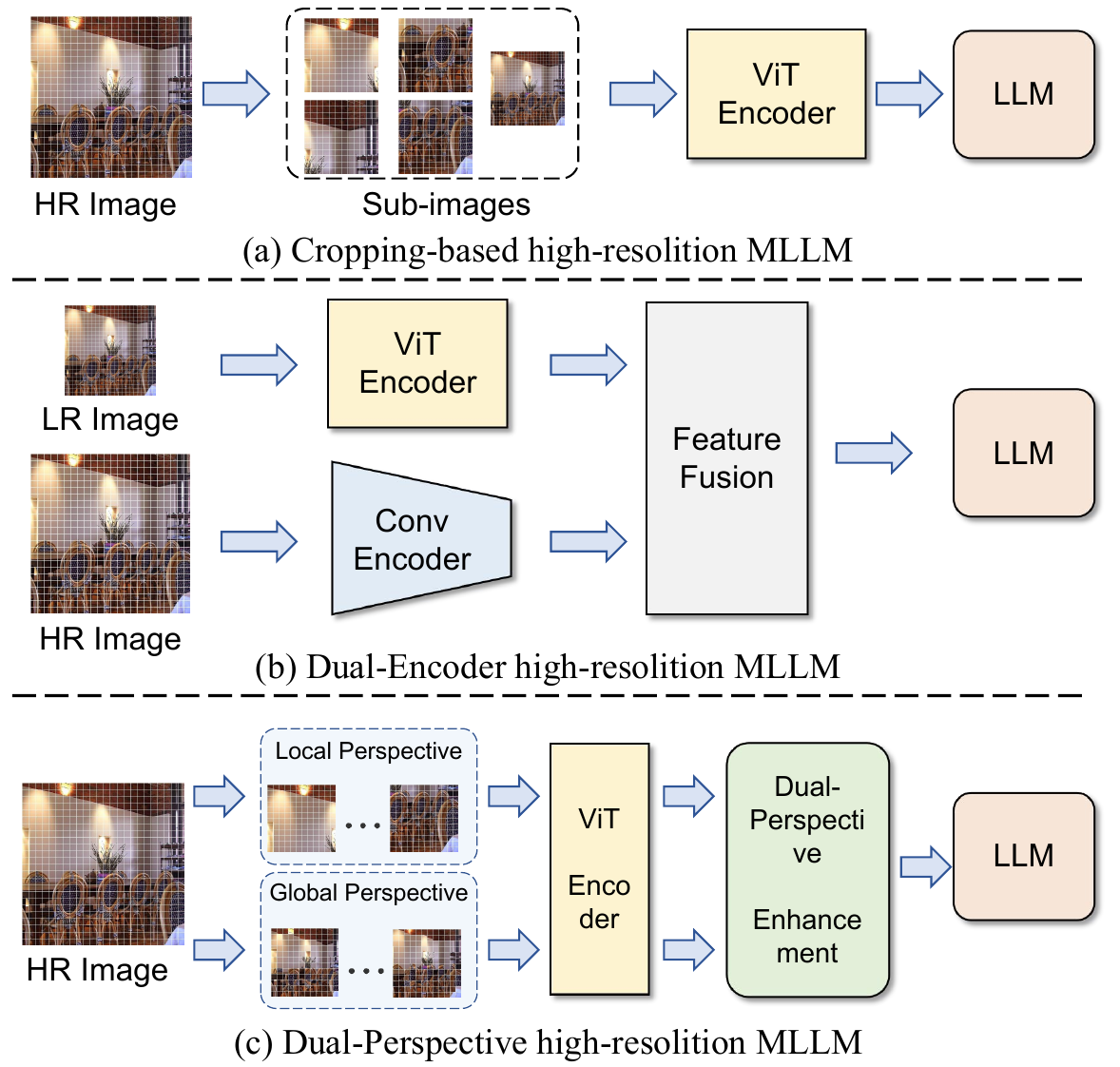}
\caption{{Comparison between existing high-resolution MLLMs and \modelname.} LR and HR abbreviate low-resolution and high-resolution, respectively. Zoom in for optimal viewing.
}
\label{fig:intro}
\end{figure}

Extensive research has highlighted the critical importance of high-resolution imagery in computer vision, particularly for tasks requiring precise image perception, such as object detection~\cite{liu2021hrdnet, zeng2019towards} and segmentation~\cite{zhao2018icnet, lin2017refinenet}.
Similarly, enhancing resolution can significantly improve the visual acuity of Multimodal Large Language Models (MLLMs). High-resolution input images inherently provide enriched detail and intricate object relationships, which are essential for mitigating hallucination issues~\cite{bai2024hallucination, jiang2024hallucination, yu2024hallucidoctor} and enhancing fine-grained perception tasks~\cite{hu2024mplug, hu2023mplug}.
However, large language models (LLMs) necessitate careful control over the number of image tokens produced by the image encoder, as these tokens significantly affect inference speed and computational cost. Additionally, because the visual encoder in MLLMs is typically a Vision Transformer (ViT)~\cite{dosovitskiy2020image}, which has a computational complexity that scales quadratically with image resolution, it is crucial to limit the resolution of images fed into the ViT.
As discussed, high-resolution image perception poses significant challenges for MLLMs. Thus, achieving a balance between leveraging the advantages of high-resolution inputs and managing the practical limitations of computational resources is essential for the successful deployment of MLLMs.

To address the challenge of efficient high-resolution image perception in MLLMs, existing methods are categorized into two main approaches: \textit{cropping-based methods} and \textit{dual-encoder methods}, as depicted in Fig.~\ref{fig:intro}.
Given that the ViT encoder~\cite{dosovitskiy2020image} is pretrained on low-resolution images and considering its quadratic complexity relationship with image resolution, cropping-based methods~\cite{dong2024internlm,xu2024llava,li2024monkey,lin2023sphinx} partition a high-resolution image into several sub-images. These sub-images are independently processed by the ViT encoder to extract their visual features, as illustrated in Fig.~\ref{fig:intro}(a).
However, this independent cropping and encoding approach fails to adequately model the interrelationships between the sub-images. Research~\cite{yao2018exploring,zhao20213dvg,lu2018r,cadene2019murel} underscores that understanding object relationships is essential for comprehensive image interpretation.
Recognizing a linear relationship between the complexity of convolutional neural networks (CNNs)~\cite{liu2022convnet,li2021survey} and image resolution, some researchers~\cite{li2024mini,luo2024feast} have proposed a dual-encoder approach. This method leverages a pretrained ConvNeXt~\cite{liu2022convnet} encoder to supplement the ViT encoder for high-resolution image perception, illustrated in Fig.~\ref{fig:intro}(b).
However, the dual-encoder method requires additional pretrained convolutional neural networks, necessitating extensive computational resources and large-scale datasets~\cite{schuhmann2021laion,schuhmann2022laion}, often demanding thousands of GPU hours. 

In this paper, we introduce \modelname, a highly effective and efficient framework designed to enhance input image resolution within multimodal large language models (MLLMs), as illustrated in Fig.~\ref{fig:intro}(c).
The framework incorporates two innovative designs that significantly improve image resolution handling.
Firstly, we propose \moduleOneBig~(\moduleOneBigShort), a sophisticated cropping strategy that partitions high-resolution images into sub-images from both local and global perspectives. From the local perspective, sub-images maintain continuous, detailed information, capturing essential details from various regions of the original image. From the global perspective, sub-images aggregate global information, albeit with less detail, as each patch in these sub-images is cropped from the original image following a specific stride pattern. This method approach ensures that \moduleOneBigShort surpasses previous cropping methods by preserving the integrity of both global and local information at the cropping stage.
Secondly, we introduce \moduleTwoBig~(\moduleTwoBigShort) to facilitate interaction between local and global features. While a straightforward approach would involve cross-attention between these features, the quadratic increase in token number due to high-resolution images often results in out-of-memory issues. To address this, our module applies a more resource-efficient strategy: it concatenates global-perspective sub-image features back into the original image's shape based on 2D priors. These concatenated global features are then re-cropped into multiple sub-images from a local perspective. Each newly generated sub-image is matched with its corresponding local perspective sub-image, and cross-attention is performed to enrich the global features with enhanced local details. Additionally, symmetric operations are applied to local-perspective sub-images to bolster global information.
Building upon \moduleOneBigShort and \moduleTwoBigShort, we propose a new high-resolution MLLM, namely \modelname.
Experimental results overwhelmingly demonstrate that these innovative designs not only enhance the handling of high-resolution images within MLLMs but also significantly optimize computational efficiency, establishing \modelname as a compelling solution to advance the field.

In summary, our contributions are three-fold:

\begin{itemize}
    \item We propose a novel \moduleOneBig (\moduleOneBigShort), which integrates both global and local perspectives when cropping high-resolution images into sub-images. This enhances the model's ability to capture detailed and contextual information.
    \item We introduce \moduleTwoBig (\moduleTwoBigShort), an effective and efficient module for fusing dual-perspective features, resulting in dual-enhanced features that significantly improve performance.
    \item Based on these two novel modules, we develop \modelname, a powerful MLLM that outperforms existing models on multiple benchmarks, demonstrating the effectiveness of our approach.
\end{itemize}

\section{Related Work}

\subsection{Large Language Models (LLMs)}
In the early stages of natural language processing (NLP) advancements, models like GPT-2~\cite{radford2019language} and BERT~\cite{devlin2018bert}, pretrained on web-scale text datasets, showcased exceptional representational capabilities. These models achieved monumental success and marked a significant breakthrough in the field of NLP.
Building on the effectiveness of the pre-training paradigm, researchers have further enhanced large language models (LLMs) by increasing the amount of pre-training data and scaling up model parameters. Representative works in this domain include GPT-3~\cite{brown2020language}, PaLM~\cite{chowdhery2023palm}, and OPT~\cite{zhang2022opt}, which have each set new benchmarks for performance and capability.
Recent efforts have pivoted towards improving LLM responses to be more aligned with human preferences by incorporating human instructions and feedback. Notable examples include InstructGPT~\cite{ouyang2022training}, ChatGPT~\cite{openai2022chatgpt}, and GPT-4~\cite{achiam2023gpt}, which demonstrate strong perceptual and reasoning abilities in human conversations. These models have advanced the state of conversational AI, making interactions more intuitive and human-like.
Additionally, the open-source LLaMA series~\cite{touvron2023llama,touvron2023llama2,touvron2023llama3} represent a significant contribution to the field. To further enhance the human interaction capabilities of LLaMA, researchers have developed Alpaca~\cite{alpaca}, Vicuna~\cite{vicuna2023}, and MPT~\cite{mosaicml2023mpt}, which fine-tune the LLaMA model using additional high-quality instruction data.
Recognizing the importance of aligning models with human intentions and preferences, some researchers~\cite{achiam2023gpt,touvron2023llama2,anthropic2023claude} have incorporated Reinforcement Learning from Human Feedback (RLHF)~\cite{schulman2017proximal,rafailov2024direct} into the training process. This approach ensures that models not only respond accurately but also in ways that are aligned with human values and requirements, thereby significantly enhancing the user experience and reliability of AI systems.

\subsection{Multimodal Large Language Models (MLLMs)}
Multimodal Large Language Models (MLLMs) are designed to extend the capabilities of traditional large language models (LLMs) by incorporating both textual and visual understanding, thereby enhancing their ability to interpret visual information and provide contextually rich responses.
MLLMs~\cite{zhao2024mg, zhu2023minigpt, lu2024deepseek} generally comprise three core components: a vision encoder, a connector, and an LLM.
The vision encoder acts as the "eyes" of the model, enabling it to perceive and analyze visual content. This encoder can utilize various structures, such as Vision Transformer (ViT)~\cite{dosovitskiy2020image} or ConvNeXt~\cite{liu2022convnet}, and can be pretrained using different methodologies, including self-supervised learning~\cite{caron2021emerging, oquab2023dinov2} or supervised learning~\cite{radford2021learning}. Most MLLMs employ CLIP-ViT, which is pre-trained on extensive image-text pairs, as the vision encoder to extract visual features effectively.
The connector in MLLMs is responsible for transforming these visual features into the textual domain, facilitating seamless integration with the LLM. There are three prevalent types of projectors:
\textit{1) Cross-attention-based methods}: Models like Flamingo~\cite{alayrac2022flamingo} and CogVLM~\cite{wang2023cogvlm} utilize cross-attention mechanisms to interweave visual and textual tokens within the LLM, effectively merging the two modalities.
\textit{2) Query-based methods}: Approaches such as Blip-2~\cite{li2023blip}, Instruct-Blip~\cite{dai2024instructblip}, and Qwen-VL~\cite{bai2023qwen} employ learnable queries to extract visual features using transformer-like architectures. These queries are then concatenated with text tokens, and the combined tokens are fed into the LLM.
\textit{3) Projection-based methods}: Techniques like LLaVA~\cite{liu2024visual, liu2024improved}, Mini-GPT4~\cite{zhu2023minigpt}, and DeepSeek-VL~\cite{lu2024deepseek} leverage a linear layer or a multi-layer perceptron (MLP) to project visual tokens into the textual domain directly, subsequently feeding the mixed tokens into the LLM.
The LLM, serving as the "brain" of the MLLM, interprets and processes the combined text and image information, delivering coherent and contextually appropriate responses. The range of LLMs available for integration is extensive, including models like LLaMA~\cite{touvron2023llama, touvron2023llama2, touvron2023llama3}, Qwen~\cite{bai2023qwen}, DeepSeek~\cite{bi2024deepseek}, and Yi~\cite{young2024yi}.
Through the synergistic combination of these sophisticated components, MLLMs significantly enhance the capabilities of traditional LLMs, enabling them to seamlessly integrate and process multiple modalities.

\subsection{High-resolution MLLMs}
High-resolution images offer significant advantages for Multimodal Large Language Models (MLLMs) by enabling the capture of detailed object information and complex relationships between objects in images.
However, directly inputting high-resolution images into the vision encoder results in prohibitive computational expenses, primarily due to the quadratic complexity associated with the Transformer architecture~\cite{vaswani2017attention} and the substantial increase in the number of visual tokens.
To mitigate this issue, existing high-resolution MLLMs can be categorized into two primary types: \textit{Cropping-based methods} and \textit{Dual-Encoder methods}, as illustrated in Fig.~\ref{fig:intro}(a) and Fig.~\ref{fig:intro}(b).
Cropping-based methods~\cite{liu2024llava,ye2023ureader,lin2023sphinx,dong2024internlm} partition an image into multiple non-overlapping patches and feed each patch into the vision encoder separately, thereby obtaining visual features for local regions. To ensure that each patch maintains an aspect ratio close to 1:1, LLaVA-UHD~\cite{xu2024llava} introduces various patching strategies during the crop operation. Furthermore, to individually model each patch’s information, Monkey~\cite{li2024monkey} employs LoRA~\cite{hu2021lora} to fine-tune the vision encoder for each specific patch.
Despite their benefits, cropping-based methods can disrupt the global coherence of image information by segmenting a complete image into isolated sub-images. As a result, some researchers have proposed dual-encoder methods to maintain the integrity of global information.
Dual-encoder methods leverage an auxiliary visual encoder to enhance high-resolution image understanding without significantly increasing the number of visual tokens. For instance, Vary~\cite{wei2023vary} and Deepseek-VL~\cite{lu2024deepseek} utilize the Segment Anything Model (SAM)~\cite{kirillov2023segment} within a high-resolution vision encoder to better capture high-resolution information. Meanwhile, MiniGemini~\cite{li2024mini} and LLaVA-HR~\cite{luo2024feast} employ ConvNeXt~\cite{liu2022convnet}, pretrained on the massive LAION2B dataset~\cite{schuhmann2021laion}, to augment the visual features extracted by the Vision Transformer (ViT).
However, dual-encoder methods necessitate an additional pretrained vision encoder to process high-resolution images. Both SAM, pretrained on the SA-1B dataset, and ConvNeXt, pretrained on the LAION-2B dataset, require extensive computational resources, amounting to tens of thousands of GPU hours, which can be cost-prohibitive.
In this paper, we introduce \modelname, a novel framework that addresses these challenges by integrating an innovative \moduleOneBig and a new \moduleTwoBig to enhance the preservation of high-resolution global information. Our approach ensures not only the efficiency of computational resources but also the comprehensive capture of both local and global image details, thereby advancing the capabilities of high-resolution MLLMs.

\section{Preliminary}

A multimodal large language model (MLLM) is an advanced AI system designed to handle and integrate both visual and textual data effectively. It typically consists of three primary components: an image encoder $\mathcal{F}_I(\cdot)$, a connector $\mathcal{F}_C(\cdot)$, and a well-pretrained large language model (LLM) $\mathcal{F}_L(\cdot)$. The image encoder processes the input image $I \in \mathbb{R}^{H \times W \times 3}$, where $H$ and $W$ represent the height and width of the image, respectively, and the three denotes the RGB color channels. This encoder extracts high-dimensional visual features from the image. The connector maps these visual features into a format that the LLM can interpret, effectively serving as a bridge between the visual and textual domains. The LLM, pretrained on a vast amount of textual data, processes the integrated visual and textual data to generate coherent and contextually relevant responses.

The input to the MLLM typically includes an image $I$ and a corresponding instruction text $T_{\text{ins}} \in \mathbb{R}^{L}$, where $L$ is the number of tokens in the instruction. Initially, the image $I$ is processed by the image encoder $\mathcal{F}_I(I)$ to extract visual features. Concurrently, the instruction $T_{\text{ins}}$ is tokenized using the tokenizer $\mathcal{F}_T$ of the LLM to convert the text into a series of tokens. The extracted visual features are then flattened and projected into visual tokens. The connector $\mathcal{F}_C$ converts these visual tokens into a format compatible with the LLM. Subsequently, the visual tokens and the textual tokens are concatenated along the spatial dimension and fed into the LLM $\mathcal{F}_L$.

The LLM decodes the combined visual and textual tokens to generate a response token-by-token. Mathematically, this decoding process can be formulated as:
\begin{equation}
\begin{split}
p(R_t \mid &I, T_{\text{ins}}, R_{0:t-1}) = \\ 
&\mathcal{F}_L
\left(
R_t \mid \mathcal{F}_C\big(\mathcal{F}_I(I)\big), \mathcal{F}_T\big(T_{\text{ins}}\big), \mathcal{F}_T\big(R_{0:t-1}\big)
\right).
\label{eq:decode}
\end{split}
\end{equation}

Here, $p(R_t \mid I, T_{\text{ins}}, R_{0:t-1})$ represents the probability distribution of the predicted token $R_t$ at time $t$, given the image, instruction text, and the previously generated tokens. $\mathcal{F}_T(R_{0:t-1})$ denotes the tokenized form of the response generated up to token $t-1$, and $R_t$ is the $t$-th token of the generated response.

\section{Methods}
In this section, we commence by providing a comprehensive overview of the proposed \modelname framework in Sec.~\ref{sec:overview}, highlighting its innovative structure and capabilities. Next, we delve into the specifics of the two critical components: \moduleOneBig and \moduleTwoBig, thoroughly examining their intricate functionalities in Sec.~\ref{sec:moduleone} and Sec.~\ref{sec:moduletwo}, respectively. 

\begin{figure*}[]
\centering
\includegraphics[width=2.0\columnwidth]{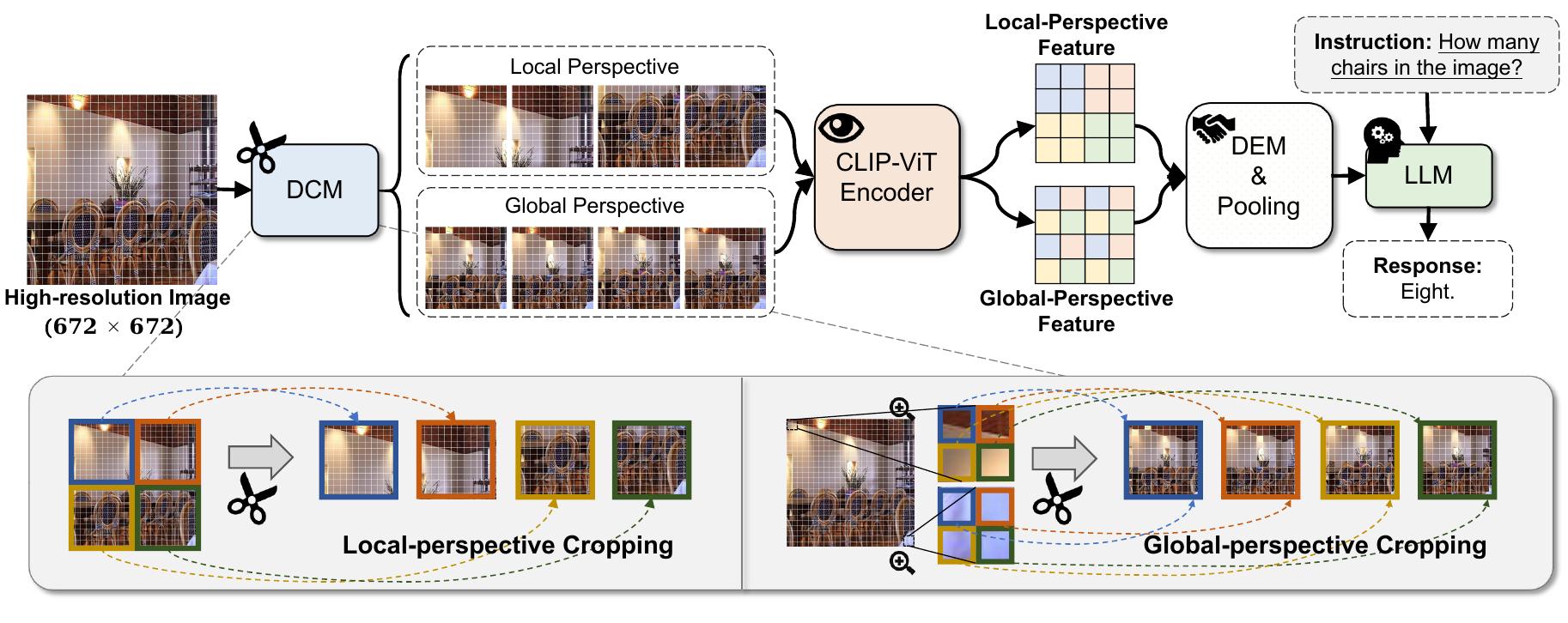}
\caption{Overview of the proposed \modelname framework. 
To address the limitations of processing high-resolution images directly with the pretrained CLIP-ViT encoder, \moduleOneBig (\moduleOneBigShort) segments the high-resolution image into sub-images from both local and global perspectives. 
Each sub-image is then individually passed through the CLIP-ViT encoder to extract distinct visual features.
These features are subsequently recombined based 2D positional priors, resulting in a comprehensive set of high-resolution local and global features.
\moduleTwoBig (\moduleTwoBigShort) is introduced to facilitate effective interaction between the local and global features.
Next, an average pooling layer is applied to reduce the number of visual tokens, enhancing computational efficiency and speeding up both training and inference processes. 
Finally, the refined visual tokens are concatenated with textual tokens of the instruction and fed into the LLM, which generates responses sequentially, token by token.
}
\label{fig:overview}
\end{figure*}

\subsection{Overview}
\label{sec:overview}
Fig.~\ref{fig:overview} depicts the comprehensive pipeline of the proposed \modelname framework. Due to the pretrained ViT-CLIP encoder's limitation in processing high-resolution images—given its quadratic complexity characteristics—directly feeding high-resolution images into it is computationally prohibitive.
To address this challenge, we propose \moduleOneBig (\moduleOneBigShort) $\mathcal{F}_{DCM}(\cdot)$, which partitions high-resolution images into several sub-images from both global and local perspectives, using the resolution defined by the pretrained vision encoder. Mathematically, this operation can be described as:
\begin{equation}
\begin{split}
[I^{loc}_{1}, I^{loc}_{2}, \cdots, I^{loc}_{N}; I^{glo}_{1}, I^{glo}_{2}, \cdots, I^{glo}_{N}] = \mathcal{F}_{DCM}(I),
\label{eq:dcm}
\end{split}
\end{equation}
where $[I^{loc}_{1}, I^{loc}_{2}, \cdots, I^{loc}_{N}]$ and $[I^{glo}_{1}, I^{glo}_{2}, \cdots, I^{glo}_{N}]$ represent the sub-images from local and global perspectives, respectively. Here, $N$ denotes the number of sub-images from each perspective, and $I \in \mathbb{R}^{W_h \times H_h \times 3}$ is the high-resolution input image.
Next, each local and global sub-image is separately fed into the pretrained vision encoder:
\begin{equation}
\mathbf{F}^{loc}_i = \mathcal{F}_I(I^{loc}_i),
\label{eq:encode1}
\end{equation}
\begin{equation}
\mathbf{F}^{glo}_i = \mathcal{F}_I(I^{glo}_i),
\label{eq:encode2}
\end{equation}
where $\mathbf{F}^{loc}_i \in \mathbb{R}^{w_{l} \times h_{l} \times d}$ and $\mathbf{F}^{glo}_i \in \mathbb{R}^{w_{l} \times h_{l} \times d}$ are the visual features of the $i$-th local and global sub-images, respectively. Here, $w_{l} \times h_{l}$ denotes the number of visual tokens per sub-image, and $d$ is the channel dimension of the visual features.
These local and global sub-image features are then recombined using 2D positional prior information to form high-resolution image features:
\begin{equation}
\mathbf{F}^{loc} = \mathcal{F}_{loc}(\mathbf{F}^{loc}_1, \mathbf{F}^{loc}_2, \cdots, \mathbf{F}^{loc}_N),
\label{eq:recombine1}
\end{equation}
\begin{equation}
\mathbf{F}^{glo} = \mathcal{F}_{glo}(\mathbf{F}^{glo}_1, \mathbf{F}^{glo}_2, \cdots, \mathbf{F}^{glo}_N),
\label{eq:recombine2}
\end{equation}
where $\mathcal{F}_{loc}(\cdot)$ and $\mathcal{F}_{glo}(\cdot)$ are the recombination functions based on local and global positional information. The resulting features $\mathbf{F}^{loc} \in \mathbb{R}^{w_{h} \times h_{h} \times d}$ and $\mathbf{F}^{glo} \in \mathbb{R}^{w_{h} \times h_{h} \times d}$ represent the high-resolution visual features from local and global perspectives, respectively, where $w_{h} \times h_{h}$ is the number of visual tokens in the recombined high-resolution images.
To facilitate efficient interaction between the local and global features, the proposed \moduleTwoBig (\moduleTwoBigShort) $\mathcal{F}_{DEM}(\cdot)$ is employed. This module ensures a robust exchange of information between local and global features, resulting in dual-enhanced features:
\begin{equation}
\mathbf{F}^{dual} = \mathcal{F}_{pool}(\mathcal{F}_{DEM}(\mathbf{F}^{loc}, \mathbf{F}^{glo})),
\label{eq:dem}
\end{equation}
where $\mathcal{F}_{pool}(\cdot)$ is the average pooling function used to reduce the number of visual tokens, thereby accelerating training and inference speeds while minimizing computational overhead. The resulting $\mathbf{F}^{dual} \in \mathbb{R}^{w_l \times h_l \times d}$ represents the visual features enhanced from dual perspectives.\footnote{ \(W_h \times H_h\) and \(W_l \times H_l\) denote the resolutions of the high-resolution and low-resolution images, respectively. Furthermore, \(w_h \times h_h\) and \(w_l \times h_l\) represent the resolutions of the high-resolution and low-resolution visual features, respectively.}
Finally, the connector $\mathcal{F}_C(\cdot)$ projects the dual-enhanced visual features to obtain visual tokens that align with the textual features. The instruction $T_{ins}$ is tokenized by the tokenizer $\mathcal{F}_T(\cdot)$, converting it into a sequence of tokens. These visual tokens and textual tokens are then concatenated along the spatial dimension and fed into the pretrained LLM to generate the response:
\begin{equation}
R = \mathcal{F}_L(\mathcal{F}_C(\mathbf{F}^{dual}), \mathcal{F}_T(T_{ins})),
\label{eq:decode}
\end{equation}
where $R \in \mathbb{R}^{L_{res}}$ represents the response generated by the LLM. Here, $L_{res}$ is the length of the generated response in tokens.

\subsection{\moduleOneBig}
\label{sec:moduleone}

\moduleOneBig (\moduleOneBigShort) is designed to effectively partition the high-resolution image \(I \in \mathbb{R}^{W_h \times H_h \times 3}\) into multiple sub-images \(I_i \in \mathbb{R}^{W_l \times H_l \times 3}\), where \(W_l \times H_l\) corresponds to the resolution utilized by the vision encoder during pretraining. For instance, in the case of the CLIP-ViT-large-patch14-336 encoder, \(W_l = H_l = 336\). The primary objective of \moduleOneBigShort is to perform cropping from both local and global perspectives to capture fine-grained details and broader contextual information, respectively.

In the sections that follow, we will elaborate on the methodologies employed for cropping high-resolution images from both perspectives, ensuring that the integrity and essential characteristics of the original image are preserved.

\subsubsection{Local-perspective Cropping}
The local-perspective cropping technique is designed to systematically extract smaller regions from the high-resolution image while preserving detailed and continuous visual information. This approach ensures that each sub-image retains the high-resolution details necessary for accurate analysis and representation. By strategically segmenting the high-resolution image into localized sub-images, \moduleOneBigShort effectively maintains the intricate details of the original image.

Given a high-resolution image \(I \in \mathbb{R}^{W_h \times H_h \times 3}\), the first step is to determine the relationship between the dimensions of the high-resolution image (\(W_h, H_h\)) and the dimensions expected by the pretrained vision encoder (\(W_l, H_l\)). This relationship is quantified as follows:
\begin{equation}
n_W = \left\lfloor \frac{W_h}{W_l} \right\rfloor,
\label{eq:nw}
\end{equation}
\begin{equation}
n_H = \left\lfloor \frac{H_h}{H_l} \right\rfloor,
\label{eq:nh}
\end{equation}
where \(\left\lfloor \cdot \right\rfloor\) represents the floor function, which rounds down to the nearest integer. \(n_W\) and \(n_H\) correspond to the number of local sub-images along the width and height of the high-resolution image, respectively.

Thus, the high-resolution image \(I\) is divided into \(n_W \times n_H\) local sub-images \(I^{loc}_{i}\), each sized \(W_l \times H_l\). Specifically, for each sub-image \(I^{loc}_{i}\), where \(i \in [0, n_H \times n_W - 1]\), the following formulas determine the row and column positions:
\begin{equation}
\text{row} = \left\lfloor \frac{i}{n_W} \right\rfloor,
\end{equation}
\begin{equation}
\text{col} = i \bmod n_W,
\end{equation}
where \(\bmod\) denotes the modulo operation.
The bounding box for each local sub-image \(I^{loc}_{i}\) can be described by:
\begin{equation}
\begin{split}
I^{loc}_{i} = I\big[\text{row} \cdot H_l : (\text{row} + 1) \cdot H_l, \\
\, \text{col} \cdot W_l : (\text{col} + 1) \cdot W_l, \, : \big].
\label{eq:local_box}
\end{split}
\end{equation}
This process ensures that each local sub-image \(I^{loc}_{i}\) contains continuous and detailed visual information from the high-resolution image, thereby preserving the integrity and quality of the image for localized analysis.

By using the local-perspective cropping technique, \moduleOneBigShort maintains high-fidelity representation in each sub-image, enabling robust feature extraction and analysis. This approach ensures that finer details are not lost, providing a comprehensive understanding of localized regions without compromising on resolution or detail.

\subsubsection{Global-perspective Cropping}

Conversely, global-perspective cropping aims to capture broader contextual information from high-resolution images to preserve the spatial relationships between objects. This approach ensures that our model retains an understanding of both micro and macro-level details within the extracted sub-images, facilitating the integration of comprehensive global contextual information.

Given the high resolution \(W_h \times H_h\) of the input image and the low resolution \(W_l \times H_l\) used by the pretrained vision encoder, the number of global sub-images \(n_W \times n_H\)  is computed similarly to the local perspective, as shown in Equ.~\eqref{eq:nw} and \eqref{eq:nh}:

\begin{equation}
n_W = \left\lfloor \frac{W_h}{W_l} \right\rfloor,
\label{eq:nw}
\end{equation}

\begin{equation}
n_H = \left\lfloor \frac{H_h}{H_l} \right\rfloor.
\label{eq:nh}
\end{equation}

For the \(i\)-th row and \(j\)-th column of sub-images, the pixel indices set \({I}_{ij}\) is defined as follows:

\begin{equation}
I_{i j}^{g l o}=\left\{I_{(x, y)} \left\lvert\, \begin{array}{l}
x=j+m \cdot n_{W}, \\
y=i+n \cdot n_{H}, \\
0 \leq m<\left\lfloor\frac{W_{h}}{n_{W}}\right\rfloor, m\in \mathbb{N}\\
0 \leq n<\left\lfloor\frac{H_{h}}{n_{H}}\right\rfloor, n\in \mathbb{N}
\end{array}\right.\right\},
\end{equation}
where \((x, y)\) corresponds to the pixel indices within the high-resolution image. $\mathbb{N}$ represents the set of natural numbers. Thus, each pixel \((u, v)\) in the \(I^{glo}_{ij}\) sub-image can be mapped back to the high-resolution image \(I\) as:
\begin{equation}
I^{glo}_{ij}(u, v) = I\left( j + u \cdot n_W, \, i + v \cdot n_H \right).
\end{equation}

This module effectively partitions the high-resolution image into sub-images that encapsulate the global perspective by interleaving pixels from different regions. Consequently, it enables the model to maintain a coherent global context alongside the detailed local information captured by global-perspective cropping.

\begin{figure}[]
\centering
\includegraphics[width=1.0\columnwidth]{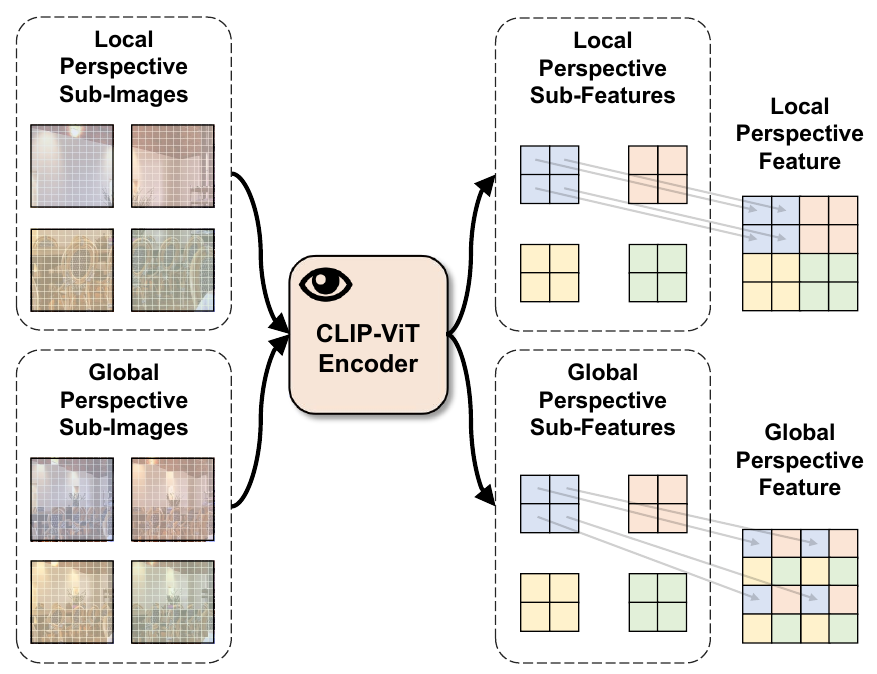}
\caption{Illustration of the integration of local and global perspective sub-features using two-dimensional positional priors. This approach ensures a seamless combination of detailed local information with broader contextual insights, maintaining spatial coherence and enhancing the overall representation of the high-resolution image.}
\label{fig:combine}
\end{figure}

\subsection{\moduleTwoBig}
\label{sec:moduletwo}

In Sec.~\ref{sec:moduleone}, we meticulously detailed the process of segmenting a high-resolution image into multiple sub-images from both local and global perspectives. Building on this foundational step, we now delve into the methods for extracting sub-features from these sub-images and their subsequent integration into high-resolution local-perspective and global-perspective features, as discussed in Sec.~\ref{sec:subfeature}. 
Next, we elucidate the sophisticated techniques employed for enhancing these features—specifically, the global-perspective enhancement in Sec.~\ref{sec:global_enhance} and the local-perspective enhancement in Sec.~\ref{sec:local_enhance}. These enhancement techniques are designed to amplify the specific details and contextual richness of their respective features, ensuring a thorough and nuanced capture of both fine-grained and broad-spectrum information.
Finally, in Sec.~\ref{sec:fusion}, we detail the fusion process where we meticulously integrate the locally-enhanced global features with the globally-enhanced local features. This fusion results in a dual-enhanced feature, which effectively combines detailed local information with comprehensive global insights. This robust integration provides an enriched and holistic representation of the high-resolution image, significantly enhancing the model's performance in various high-level visual analysis.

\subsubsection{Sub-features Combination}
\label{sec:subfeature}

Upon obtaining the local-perspective sub-images \([I^{loc}_1, I^{loc}_2, \cdots, I^{loc}_N]\) and global-perspective sub-images \([I^{glo}_1, I^{glo}_2, \cdots, I^{glo}_N]\) from \moduleOneBigShort, we proceed to extract their corresponding features using Equ.~\eqref{eq:encode1} and Equ.\eqref{eq:encode2}. This extraction yields local-perspective sub-features \([\mathbf{F}^{loc}_1, \mathbf{F}^{loc}_2, \cdots, \mathbf{F}^{loc}_N]\) and global-perspective sub-features \([\mathbf{F}^{glo}_1, \mathbf{F}^{glo}_2, \cdots, \mathbf{F}^{glo}_N]\).
As illustrated in Fig.~\ref{fig:combine}, the sub-features from both perspectives are then systematically recombined. Specifically, the local-perspective sub-features \([\mathbf{F}^{loc}_1, \mathbf{F}^{loc}_2, \cdots, \mathbf{F}^{loc}_N]\) are aggregated to construct the comprehensive local-perspective feature \(\mathbf{F}^{loc} \in \mathbb{R}^{w_h \times h_h \times d}\). Likewise, the global-perspective sub-features \([\mathbf{F}^{glo}_1, \mathbf{F}^{glo}_2, \cdots, \mathbf{F}^{glo}_N]\) are amalgamated to form the global-perspective feature \(\mathbf{F}^{glo} \in \mathbb{R}^{w_h \times h_h \times d}\).
This meticulous combination process ensures that the local-perspective feature effectively encapsulates fine-grained details, while the global-perspective feature maintains a coherent understanding of the broader contextual information.

\subsubsection{Global-Perspective Enhancement}
\label{sec:global_enhance}

\begin{figure}[]
\centering
\includegraphics[width=1.0\columnwidth]{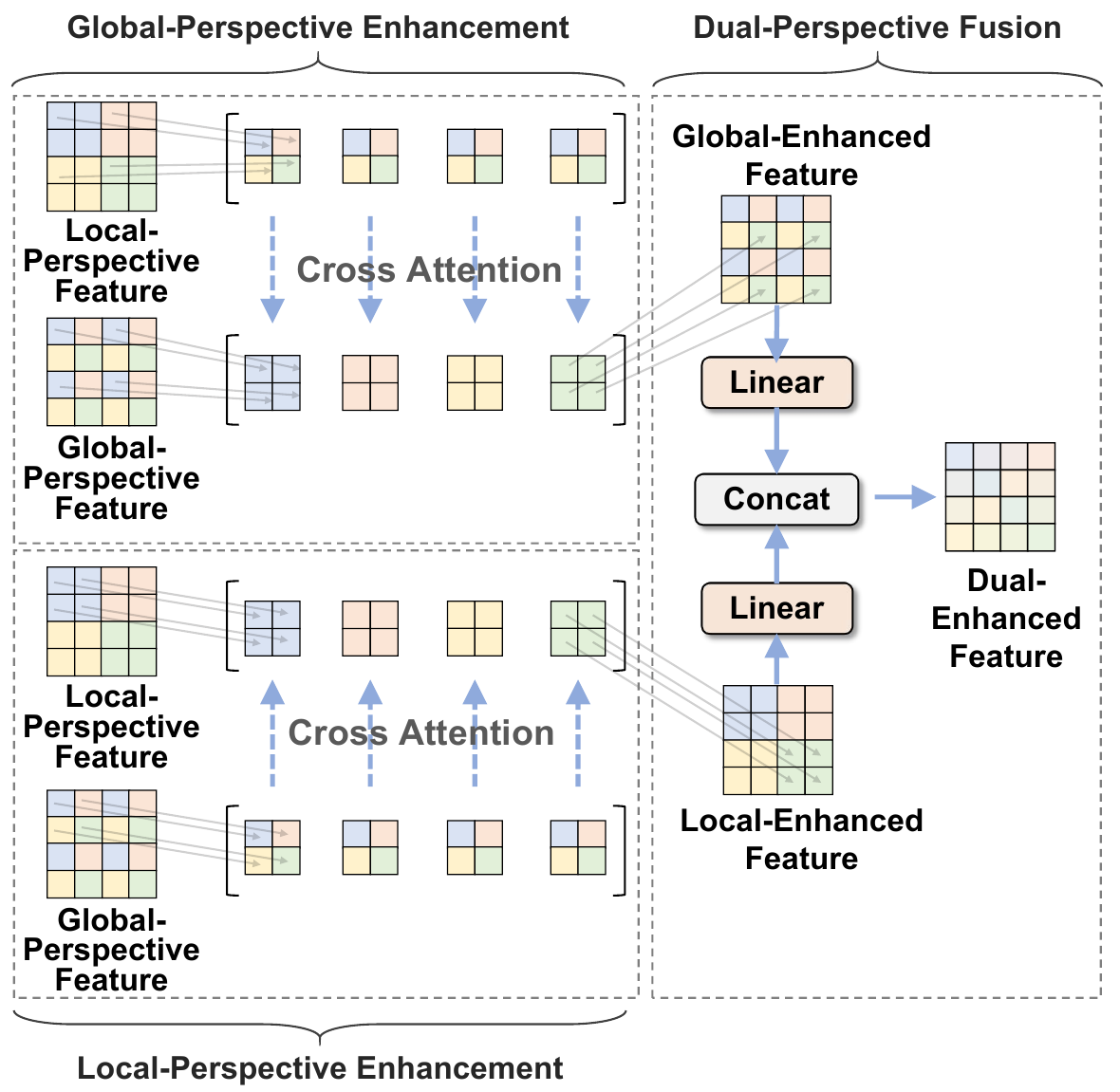}
\caption{Illustration of the proposed \moduleTwoBig (\moduleTwoBigShort), highlighting its innovative approach to efficiently integrating and enhancing local and global sub-features for superior image understanding.}
\label{fig:dem}
\end{figure}

After obtaining the local-perspective feature $\mathbf{F}^{loc} \in \mathbb{R}^{w_h \times w_l \times d}$ and global-perspective feature $\mathbf{F}^{glo} \in \mathbb{R}^{w_h \times w_l \times d}$, the next crucial step involves interacting and fusing these two feature sets to derive more robust and comprehensive features. 
A straightforward approach would be to leverage cross-attention between $\mathbf{F}^{loc}$ and $\mathbf{F}^{glo}$ to enable interaction between local and global information. However, applying cross-attention directly to such high-resolution features can lead to significant out-of-memory issues during training. To circumvent this, we propose a novel dual-perspective enhancement method that interacts with local and global features in a more efficient manner, mitigating these computational challenges.

Specifically, for global-perspective enhancement, as illustrated in Fig.~\ref{fig:dem}, we first crop both the local-perspective and global-perspective features from a global perspective. This operation can be mathematically formulated as follows:
\begin{equation}
[\mathbf{G}^{glo}_1, \mathbf{G}^{glo}_2, \cdots, \mathbf{G}^{glo}_N] = \mathcal{C}_{glo}(\mathbf{F}^{glo}),
\end{equation}
\begin{equation}
[\mathbf{L}^{glo}_1, \mathbf{L}^{glo}_2, \cdots, \mathbf{L}^{glo}_N] = \mathcal{C}_{glo}(\mathbf{F}^{loc}),
\end{equation}
where $\mathcal{C}_{glo}(\cdot)$ denotes the global-perspective cropping operation. $[\mathbf{G}^{glo}_1, \mathbf{G}^{glo}_2, \cdots, \mathbf{G}^{glo}_N]$ and $[\mathbf{L}^{glo}_1, \mathbf{L}^{glo}_2, \cdots, \mathbf{L}^{glo}_N]$ represent the global and local sub-features, respectively, resulting from this global-perspective cropping.

To infuse local sub-feature information into the global sub-features, we perform a cross-attention operation between corresponding local and global sub-features. This can be formulated as follows:
\begin{equation}
\boldsymbol{A}^{glo}_i=\operatorname{Softmax}\left(\frac{\left(\mathbf{G}^{glo}_i \cdot \boldsymbol{W}_{q}\right) \cdot\left(\mathbf{L}^{glo}_i \cdot \boldsymbol{W}_{k}\right)^{T}}{\sqrt{d}}\right),
\end{equation}
\begin{equation}
\boldsymbol{V}^{glo}_i = \boldsymbol{A}^{glo}_i \cdot \mathbf{L}_i^{glo} \cdot \mathbf{W}_v,
\end{equation}
where $\mathbf{W}_q, \mathbf{W}_k, \mathbf{W}_v \in \mathbb{R}^{d \times d}$ are learnable embedding matrices. $A^{glo}_i \in \mathbb{R}^{h_l w_l \times h_l w_l}$ are the attention map between local sub-feature and global sub-feature. $\boldsymbol{V}^{glo}_i \in \mathbb{R}^{w_l \times h_l \times d}$ are the $i$-th global-enhanced sub-feature.

Finally, the set of enhanced sub-features $[\boldsymbol{V}^{glo}_1, \boldsymbol{V}^{glo}_2, \cdots, \boldsymbol{V}^{glo}_N]$ are combined to form the globally-enhanced feature $\boldsymbol{V}^{glo} \in \mathbb{R}^{w_h \times h_h \times d}$. This procedure ensures that the resulting feature representation captures rich, multi-scale information by effectively integrating local details with the global context, thereby enhancing the overall robustness and expressiveness of the model.

\subsubsection{Local-Perspective Enhancement}
\label{sec:local_enhance} 

Similarly, Local-Perspective Enhancement aims to refine local features by integrating global contextual information. This process ensures that the local features not only retain fine-grained details but also benefit from the broader context provided by the global features.

First, the local-perspective and global-perspective features are cropped into sub-features through a local-perspective cropping operation. This is formulated as follows:

\begin{equation}
[\mathbf{G}^{loc}_1, \mathbf{G}^{loc}_2, \cdots, \mathbf{G}^{loc}_N] = \mathcal{C}_{loc}(\mathbf{F}^{glo}),
\end{equation}
\begin{equation}
[\mathbf{L}^{loc}_1, \mathbf{L}^{loc}_2, \cdots, \mathbf{L}^{loc}_N] = \mathcal{C}_{loc}(\mathbf{F}^{loc}),
\end{equation}
where $\mathcal{C}_{loc}(\cdot)$ denotes the local-perspective cropping operation. The resulting sub-features, $[\mathbf{G}^{loc}_1, \mathbf{G}^{loc}_2, \cdots, \mathbf{G}^{loc}_N]$ and $[\mathbf{L}^{loc}_1, \mathbf{L}^{loc}_2, \cdots, \mathbf{L}^{loc}_N]$, represent the global and local sub-features respectively, generated from this local-perspective cropping.

Next, we employ a cross-attention mechanism to facilitate interaction between local and global sub-features, enhancing the local features with global context. The formulation of this interaction is as follows:

\begin{equation}
\boldsymbol{A}^{loc}_i = \operatorname{Softmax}\left(\frac{\left(\mathbf{L}^{loc}_i \cdot \mathbf{W}_{q}\right) \cdot \left(\mathbf{G}^{loc}_i \cdot \mathbf{W}_{k}\right)^{T}}{\sqrt{d}}\right),
\end{equation}
\begin{equation}
\boldsymbol{V}^{loc}_i = \boldsymbol{A}^{loc}_i \cdot \mathbf{G}^{loc}_i \cdot \mathbf{W}_v,
\end{equation}
where $\mathbf{W}_q, \mathbf{W}_k, \mathbf{W}_v \in \mathbb{R}^{d \times d}$ are learnable embedding matrices used for query, key, and value transformations, respectively. The attention map $\boldsymbol{A}^{loc}_i \in \mathbb{R}^{h_l w_l \times h_l w_l}$ captures the relationships between local and global sub-features. The resulting enhanced feature $\boldsymbol{V}^{loc}_i \in \mathbb{R}^{w_l \times h_l \times d}$ is the $i$-th local-enhanced sub-feature.

Finally, the set of local-enhanced sub-features $[\boldsymbol{V}^{loc}_1, \boldsymbol{V}^{loc}_2, \cdots, \boldsymbol{V}^{loc}_N]$ are aggregated to form the local-enhanced feature $\boldsymbol{V}^{loc} \in \mathbb{R}^{w_h \times h_h \times d}$. This process ensures that the local features are enriched with complementary global contextual information, providing a more comprehensive and robust representation of the original high-resolution image. 

\subsubsection{Dual-Perspective Fusion}
\label{sec:fusion}

After obtaining the global-enhanced feature $\boldsymbol{V}^{glo}$ and the local-enhanced feature $\boldsymbol{V}^{loc}$, the next critical step involves fusing these features to create a comprehensive representation that leverages the strengths of both perspectives. To achieve this, we employ a concatenation-based method that effectively combines the global and local features.

Initially, we utilize two separate embedding layers to reduce the dimensionality of the features. This dimensionality reduction step is essential to ensure that the subsequent concatenation is computationally efficient and to highlight the most salient aspects of each feature set. The embedding operation can be described as follows:

\begin{equation}
\tilde{\boldsymbol{V}}^{glo} = \boldsymbol{V}^{glo} \mathbf{W}^{glo}, \quad \tilde{\boldsymbol{V}}^{loc} = \boldsymbol{V}^{loc} \mathbf{W}^{loc},
\end{equation}
where $\mathbf{W}^{glo}, \mathbf{W}^{loc} \in \mathbb{R}^{d \times \frac{d}{2}}$ are learnable projection matrices for the global and local features, respectively. These matrices transform the original features into a lower-dimensional space, thereby emphasizing their most critical components.

Next, we concatenate the embedded global and local features along the channel dimension to form the dual-enhanced feature. This concatenation ensures that both global contextual information and local detailed information are retained and integrated. The fusion process is formulated as follows:

\begin{equation}
\boldsymbol{V}^{dual} = [\tilde{\boldsymbol{V}}^{glo}; \tilde{\boldsymbol{V}}^{loc}],
\end{equation}
where $[\cdot;\cdot]$ denotes the concatenation operation along the channel dimension. As a result, the dual-enhanced feature $\boldsymbol{V}^{dual} \in \mathbb{R}^{w_h \times h_h \times d}$ encapsulates a comprehensive view of the input image, combining the strengths of both perspectives to produce a robust and informative representation.

\section{Experiments}

\subsection{Evaluations}

To rigorously assess the effectiveness, robustness, and versatility of the proposed \modelname, we conduct extensive evaluations across a diverse set of vision-language benchmarks. By leveraging a broad spectrum of datasets, we ensure a comprehensive evaluation of the model's capabilities across various dimensions and contexts. These benchmarks include:

\begin{itemize}
    \item \textbf{ScienceQA-img}~\cite{lu2022learn}: ScienceQA is a large-scale multimodal dataset containing 21,208 science questions from elementary and high school curricula. Each question is annotated with lectures and explanations to provide comprehensive contextual understanding, making this dataset ideal for testing the model's ability to interpret and generate accurate, context-aware responses to science-related queries.
    
    \item \textbf{OKVQA}~\cite{marino2019ok}: This open-ended visual question answering dataset requires the model to utilize external knowledge sources to answer questions accurately. It challenges the model's proficiency in integrating visual content with external textual information, providing a robust test of the model's ability to generate responses based on prior knowledge and visual understanding.
    
    \item \textbf{SEEDBench}~\cite{li2023seed}: SEEDBench is a comprehensive benchmark tailored specifically for evaluating Multimodal Large Language Models (MLLMs). It spans 12 evaluation dimensions, including comprehension of both image and video modalities. This benchmark offers a robust framework for assessing the performance of the model across a diverse array of multimodal tasks, ensuring a holistic evaluation.
    
    \item \textbf{MMBench}~\cite{liu2023mmbench}: MMBench is a multi-modality benchmark offering a comprehensive evaluation pipeline. It features a curated dataset and introduces the innovative CircularEval strategy using ChatGPT to enhance model prediction assessment. MMBench ensures a holistic evaluation of the model's multi-modal capabilities by providing a structured and thorough testing environment.
    
    \item \textbf{MMBench-CN}~\cite{liu2023mmbench}: The Chinese version of MMBench, MMBench-CN, facilitates a direct comparison of Vision-Language Model (VLM) performance in both English and Chinese contexts with verified translations. This benchmark tests the model's multi-lingual and cross-cultural adaptability, making it an essential tool for evaluating the robustness of the model in diverse linguistic settings.
    
    \item \textbf{AI2D}~\cite{kembhavi2016diagram}: AI2D is a dataset comprising illustrative diagrams aimed at research on diagram understanding and associated question answering. This benchmark challenges the model's ability to interpret and reason about structured graphical information, testing its proficiency in understanding and generating responses based on diagrammatic data.
    
    \item \textbf{LLaVA-Bench-in-the-wild}~\cite{liu2024visual}: This benchmark evaluates the capabilities of large multimodal models on real-world tasks and domains. Featuring detailed images and curated questions, LLaVA-Bench-in-the-wild tests the model's performance on diverse and challenging datasets regularly encountered in practical applications, ensuring its real-world applicability.
    
    \item \textbf{MMMU}~\cite{yue2024mmmu}: MMMU is a benchmark designed to evaluate multimodal models on college-level tasks spanning multiple disciplines. It requires advanced reasoning and subject matter expertise, thereby testing the depth of the model's understanding and its ability to handle complex, cross-disciplinary questions, making it a critical benchmark for assessing advanced cognitive capabilities.
\end{itemize}

\subsection{Implementation Details}

The vision encoder used in our implementation is the CLIP-ViT-L/14, which has demonstrated strong performance in visual tasks. The large language model (LLM) employed is LLaMA3-8B, which provides robust language understanding and generation capabilities. 
The training of \modelname is meticulously structured into two distinct stages to ensure optimal alignment and fine-tuning of the model components.

In the first stage, the pretraining phase, our primary objective is to align the features extracted by the vision encoder with the word embeddings generated by the LLM. To achieve this, we freeze both the vision encoder and the LLM during the pretraining phase. This allows us to focus on training the \moduleTwoBigShort and the projector. The model is pretrained using the CC-595k dataset~\cite{liu2024visual}, comprising a substantial collection of aligned image-text pairs, for 1 epoch. We utilize the AdamW optimizer~\cite{loshchilov2017decoupled} with a learning rate of \(1 \times 10^{-3}\) and employ a cosine learning rate schedule to smoothly adjust the learning rate during training. A global batch size of 256 is used to ensure efficient utilization of computational resources.

In the subsequent stage, the supervised fine-tuning (SFT) phase, our goal is to refine the model's performance on downstream tasks. During this phase, we freeze the vision encoder and proceed to train the \moduleTwoBigShort, the projector, and the LLM. The model is fine-tuned using the LLaVA-656K mixture dataset~\cite{liu2024improved}, which contains an extensive and diverse set of annotations to support comprehensive learning. We employ a lower learning rate of \(2 \times 10^{-5}\) and a batch size of 128 to carefully fine-tune the model parameters, ensuring precise adjustments without overfitting.

\subsection{Quantitative Analysis}

\begin{table*}[]
\caption{Comparison with State-of-the-Art (SOTA) Methods on Vision-Language Benchmarks. Our proposed methods consistently outperform existing SOTA approaches across a variety of benchmarks. The highest score is indicated in \textbf{bold}, and the second highest score is \underline{underlined}. Note that INF-LLaVA* is trained using a larger dataset, as detailed in Tab.~\ref{tab:dataset}.}
\centering
\renewcommand\arraystretch{1.2} 
\resizebox{2\columnwidth}{!}{
\begin{tabular}{l|c|cccccccc}
\toprule
\textbf{Model}                     & \textbf{Source} & \textbf{ScienceQA-img} & \textbf{OKVQA} & \textbf{SEEDBench} & \textbf{MMB}   & \textbf{MMB-CN} & \textbf{AI2D}  & \textbf{LLaVA-wild} & \textbf{MMMU} \\
\midrule
\textbf{Flamingo-80B~\cite{alayrac2022flamingo}}     & NeurIPS'22       & -                      & 50.60          & -                  & -              & -                                & -              & -                          & -                  \\
\textbf{BLIP-2~\cite{li2023blip}}           & ICML'23          & -                      & 45.90          & -                  & -              & -                                & -              & -                          & -                  \\
\textbf{InstructBLIP-7B~\cite{dai2024instructblip}}  & NeurIPS'23       & 60.50                  & -              & -                  & -              & -                                & -              & -                          & -                  \\
\textbf{InstructBLIP-13B~\cite{dai2024instructblip}} & NeurIPS'23       & 63.10                  & -              & -                  & -              & -                                & -              & -                          & -                  \\
\textbf{LLaVA~\cite{liu2024visual}}            & NeurIPS'23       & -                      & -              & 37.00              & 38.70          & 36.40                            & -              & 62.80                      & -                  \\
\textbf{IDEFICS-9B~\cite{laurenccon2024obelics}}           & NeurIPS'23          & -                      &     38.40      & -                  &        -       & -                                & -              & -                          & -                  \\
\textbf{GILL~\cite{koh2024generating}}           & NeurIPS'23          & -                      &     -      & 52.50                  &        38.20       & -                                & -              & -                          & 28.80                  \\

\textbf{CM3Leon~\cite{yu2023scaling}}   & arXiv'23         & -                      & 23.80              & -                  & -              & -                                & -              & -                          &               \\

\textbf{OpenFlamingo~\cite{awadalla2023openflamingo}}     & arXiv'23       & -                      & 37.80          & -                  & -              & -                                & -              & -                          & -                  \\

\textbf{Shikra~\cite{chen2023shikra}}           & arXiv'23          & -                      & 47.16          & -                  & 58.80              & -                                & -              & -                          & -                  \\
\textbf{MiniGPT-4~\cite{zhu2023minigpt}}         & arXiv'23          & -                      & 37.50          & -                  & -              & -                                & -              & -                          & -              \\

\textbf{Qwen-VL-Chat~\cite{bai2023qwen}}     & arXiv'23         & 68.20                  & 56.60          & 65.40              & -              & -                                & 57.7              & -                          & -                  \\

\textbf{MiniGPT-v2~\cite{chen2023minigpt}}       & arXiv'23         & -                      & 57.80          & -                  & -              & -                                & -              & -                          & -                  \\
\textbf{OtterHD-8B~\cite{li2023otterhd}}       & arXiv'23         & -                      & -              & -                  & 58.30          & -                                & -              & -                          & -                  \\
\textbf{ImageBind-LLM~\cite{han2023imagebind}}   & arXiv'23         & 51.40                      & 51.66              & -                  & -              & -                                & -              & -                          & -              \\
\textbf{ChatBridge~\cite{zhao2023chatbridge}}   & arXiv'23         & -                      & 45.20              & -                  & -              & -                                & -              & -                          & -              \\
\textbf{AnyMAL-13B~\cite{moon2023anymal}}   & arXiv'23         & 52.70                      & 33.10              & -                  & -              & -                                & -              & -                          & -              \\
\textbf{AnyMAL-70B~\cite{moon2023anymal}}   & arXiv'23         & 70.80                      & 42.60              & -                  & -              & -                                & -              & -                          & -              \\

\textbf{Emu-I~\cite{sun2023emu}}   & ICLR'24         & -                      & 49.20              & -                  & -              & -                                & -              & -                          &        -       \\

\textbf{MGIE~\cite{fu2023guiding}}   & ICLR'24         & -                      & -              & 28.80                  & 6.60              & -                                & -              & -                          & 25.60              \\
\textbf{DreamLLM~\cite{dong2023dreamllm}}   & ICLR'24         & -                      & 52.20              & -                  & 58.20              & -                                & -              & -                          &       -        \\
\textbf{mPLUG-Owl2~\cite{ye2024mplug2}}       & CVPR'24          & 68.70                  & 57.70          & 57.80              & -              & -                                & -              & -                          & -                  \\
\textbf{Monkey~\cite{li2024monkey}}           & CVPR'24          & 69.40                  & \textbf{61.30} & -                  & -              & -                                & 57.90          & -                          & -                  \\
\textbf{LLaVA1.5~\cite{liu2024improved}}         & CVPR'24         & 66.80                  & -              & 66.10              & 64.30          & 58.30                            & -              & 65.40                      & 36.40              \\

\textbf{Unified-IO 2 ~\cite{lu2024unified}}   & CVPR'24         & 78.60                      & 50.20              & -                  & -              & -                                & -              & -                          & -              \\
\textbf{Honeybee ~\cite{cha2024honeybee}}   & CVPR'24         & -                      & -              & 64.50                  & 70.10              & -                                & -              & 67.10                          & -              \\
\textbf{OneLLM~\cite{han2024onellm}}   & CVPR'24         & 63.40                      & 58.90              & 61.20                  & 60.00              & -                                & -              & -                          & -              \\
\textbf{LoRA-Sparse~\cite{song2024low}}   & CVPR'24         & 68.40                      & -              & 58.80                  & -              & -                                & -              & 63.40                          & -              \\

\textbf{Pink~\cite{xuan2024pink}}   & CVPR'24         & -                      & 59.50              & -                  & -              & -                                & -              & -                          &     -          \\
\textbf{Prompt Highlighter~\cite{zhang2024prompt}}   & CVPR'24         & -                      & -              & -                  & 69.50              & -                                & -              & -                          &     -          \\

%
\textbf{TIVE ~\cite{liu2024less}}   & arXiv'24         & 69.20                      & -              & 62.20                  & 65.80              & 57.40                                & -              & -                          &       -        \\

\textbf{GenLLaVA ~\cite{hernandez2024generative}}   & arXiv'24         & -                      & -              & 63.50                  & 65.00              & -                                & -              & -                          & 29.70              \\

\textbf{AnyGPT~\cite{zhan2024anygpt}}   & arXiv'24         & -                      & -              & 44.50                  & 36.00              & -                                & -              & -                          & 30.60              \\

\textbf{DeepStack-L-HD~\cite{meng2024deepstack}}   & arXiv'24         & -                      & -              & -                  & -              & -                                & -              & -                          & 35.60              \\

\textbf{LLaVA-Phi~\cite{zhu2024llava}}   & arXiv'24         & 68.40                      & -              & -                  & 59.80              & -                                & -              & -                          & -              \\

\textbf{AcFormer~\cite{zong2023acformer}}         & arXiv'24         & 69.40                  & -              & -                  & 68.40          & -                                & -              & -                          & -                  \\

\textbf{DeepStack-L-HD~\cite{meng2024deepstack}}   & arXiv'24         & -                      & -              & -                  & -              & -                                & -              & -                          & 35.60              \\
\textbf{ConvLLaVA~\cite{ge2024convllava}}        & arXiv'24         & -                      & -              & 70.20              & 68.70          & -                                & -              & -                          & 35.80              \\
\textbf{LLAVA-HR~\cite{luo2024feast}}         & arXiv'24         & 65.10                  & 58.90          & 64.20              & -              & -                                & -              & -                          & -                  \\

\midrule
\textbf{INF-LLaVA}        &          Ours       & \underline{75.71}         & \underline{60.67}          & \underline{70.47}     & \underline{70.35} & \underline{63.23}                   & \underline{60.59} & \underline{67.50}             & \underline{37.00}   \\
\textbf{INF-LLaVA*}        &          Ours       & \textbf{77.44}         & 57.04          & \textbf{72.65}     & \textbf{74.38} & \textbf{64.57}                   & \textbf{75.42} & \textbf{78.20}             & \textbf{37.20}   \\

\bottomrule
\end{tabular}
}
\label{tab:sota}
\end{table*}

\begin{table}[]
    \caption{Dataset Details for Training \textbf{INF-LLaVA*}, including the datasets and corresponding Q/A pairs used in the pretraining (Stage 1) and supervised fine-tuning (Stage 2) stages.}
    \centering
    \renewcommand\arraystretch{1.5} 
    \begin{tabular}{l|c|c}
        \toprule
        \textbf{Dataset} & \textbf{Q/A pairs} & \textbf{Total} \\
        \hline
        \multicolumn{3}{l}{\textit{Stage1: Pretraining}} \\
        \hline
        LAION-CCSBU~\cite{sharma2018conceptual} & 558K & \multirow{2}{*}{1.26M} \\
        \cline{1-2}
        ALLaVA-4V-Caption~\cite{chen2024allava} & 708K & \\
        \hline
        \multicolumn{3}{l}{\textit{Stage2: Supervised Fine-tuning}} \\
        \hline
        LLaVA-Instruct~\cite{liu2024improved} & 665K & \multirow{6}{*}{1.3M} \\
        \cline{1-2}
        ALLaVA-4V-Instruction~\cite{chen2024allava} & 692K & \\
        \cline{1-2}
        ShareGPT4V~\cite{chen2023sharegpt4v} & & \\
        DocVQA~\cite{tito2021document} & 25K & \\
        DVQA~\cite{kafle2018dvqa} & & \\
        AI2D~\cite{kembhavi2016diagram} & & \\
        \bottomrule
    \end{tabular}
    \label{tab:dataset}
\end{table}

As shown in Tab.~\ref{tab:sota}, we conduct a comprehensive comparison of the proposed \modelname with existing state-of-the-art (SOTA) Multimodal Large Language Models (MLLMs) across 8 widely recognized benchmarks. The results clearly demonstrate the superior performance of \modelname. Specifically, \modelname achieves an impressive score of 75.71 on the ScienceQA-img benchmark and 70.35 on the MMBench benchmark. These scores represent significant improvements over other leading MLLMs.
Notably, \modelname outperforms QWen-VL-Chat by a considerable margin. It is important to highlight that QWen-VL-Chat leverages a substantially larger amount of data, utilizing 5 billion examples for pretraining, 76.8 million examples for multi-task training, and 350,000 in-house examples for supervised fine-tuning. In stark contrast, \modelname achieves its superior results with only 595,000 examples for pretraining and 665,000 examples for fine-tuning.
Additionally, as detailed in Tab.~\ref{tab:dataset}, we trained an enhanced version of \modelname, marked as INF-LLaVA*, using a larger dataset. Our observations indicate that when fed with more data, the performance of \modelname improves even further across most benchmarks. For instance, on the LLaVA-bench-in-the-wild benchmark, INF-LLaVA* achieves an improvement of over ten points.


\subsection{Qualitative Analysis}

\subsubsection{Comparison of Different Image Resolutions}

\begin{figure*}[]
\centering
\includegraphics[width=2.0\columnwidth]{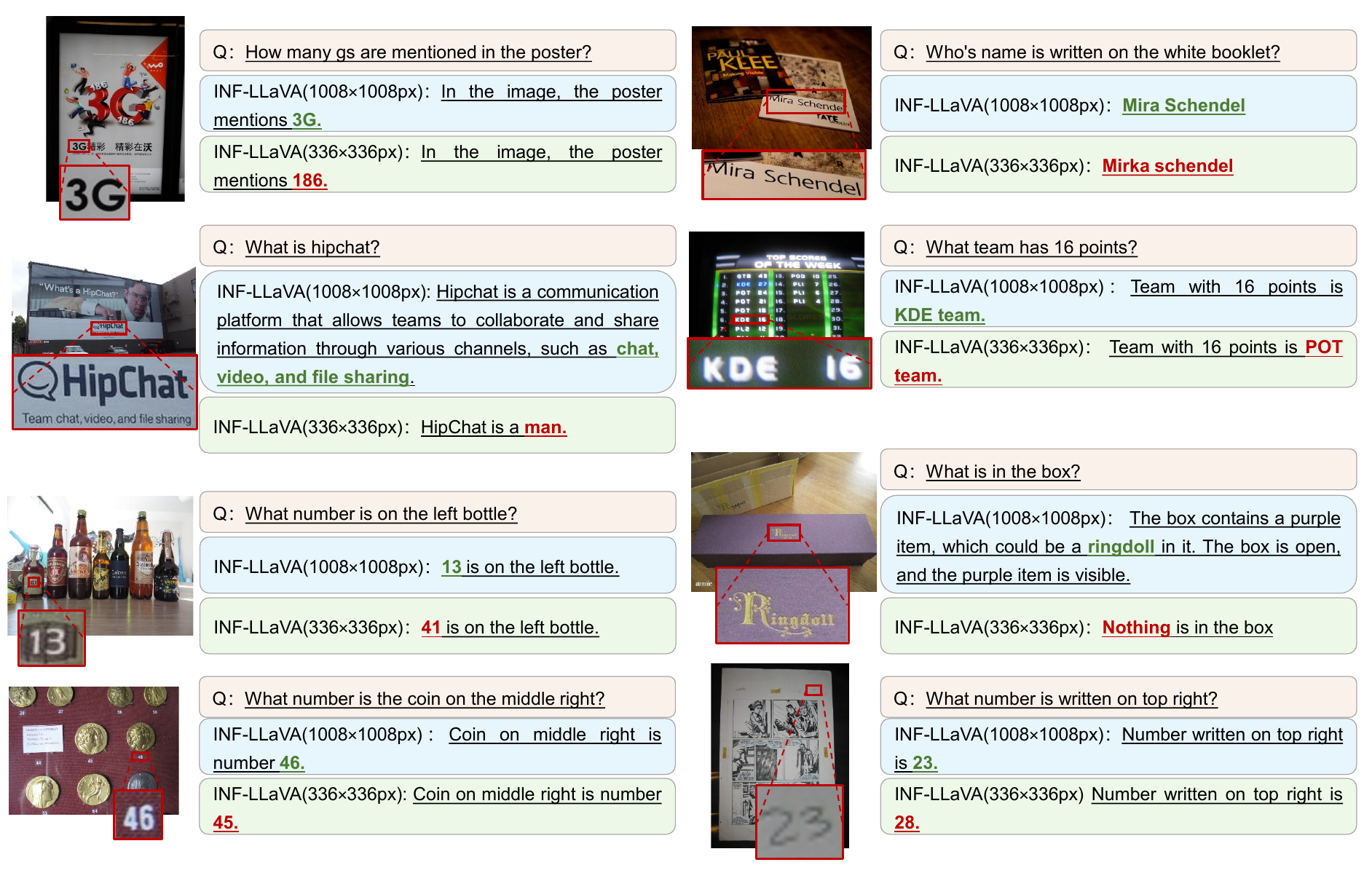}
\caption{Chat comparison using different image resolutions. Certain regions of the input high-resolution images are zoomed in for enhanced visualization.}
\label{fig:case3}
\end{figure*}

To thoroughly investigate the impact of different image resolutions on the performance of \modelname, we conducted a series of experiments using various input image resolutions. As demonstrated in Fig.~\ref{fig:case3}, leveraging high-resolution images enables \modelname to provide more accurate answers to complex questions, particularly those requiring fine-grained perception.
For instance, in the second case of the second row in Fig.~\ref{fig:case3}, the screen displays multiple team names along with their corresponding scores. Identifying the team name associated with a specific score is a challenging task. However, with an input image resolution of $1008 \times 1008$ pixels, \modelname accurately identifies the KDE team as the one that scored 16 points. In contrast, when using a lower resolution of $336 \times 336$ pixels, \modelname provides an incorrect response.
Moreover, in the second case of the third row, the task involves identifying the contents of a purple box that is closed, making it difficult to ascertain what is inside. With an input resolution of $1008 \times 1008$ pixels, \modelname successfully recognizes the "ringdoll" text on the box lid and accurately infers that the box might contain a ringdoll. Conversely, using a $336 \times 336$ pixels resolution, \modelname incorrectly concludes that the box contains no objects, demonstrating a lack of evidence-based reasoning.
These observations underscore the importance of high-resolution images in enhancing the model's ability to perceive and interpret complex visual details, thereby improving its overall accuracy and effectiveness.

\subsubsection{Comparison with LLaVA 1.5}

\begin{figure*}[]
\centering
\includegraphics[width=1.7\columnwidth]{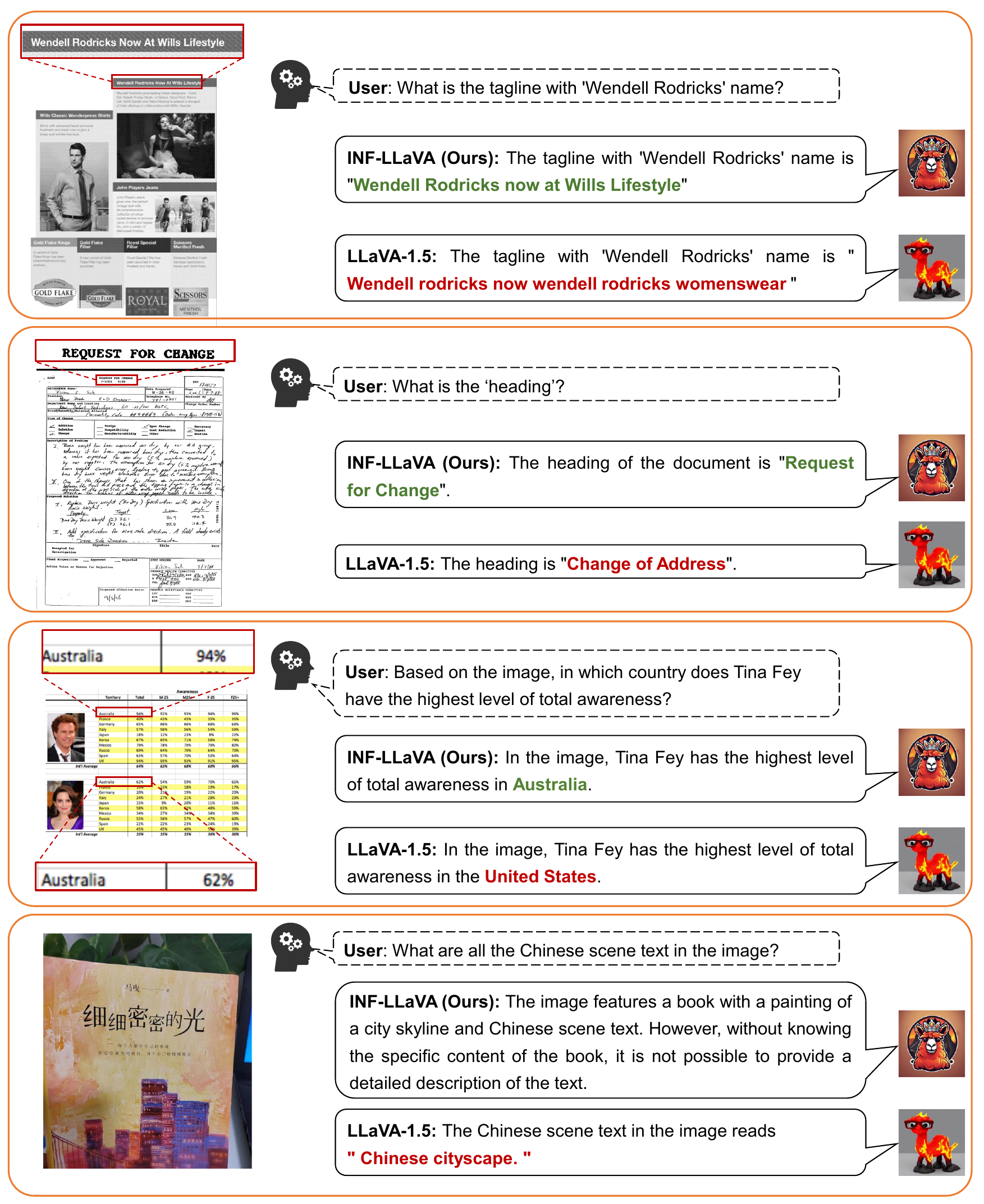}
\caption{Chat comparison between \modelname and LLaVA-1.5. Certain regions of the input high-resolution images are zoomed in for enhanced visualization.}
\label{fig:case1}
\end{figure*}

\begin{figure*}[]
\centering
\includegraphics[width=2.0\columnwidth]{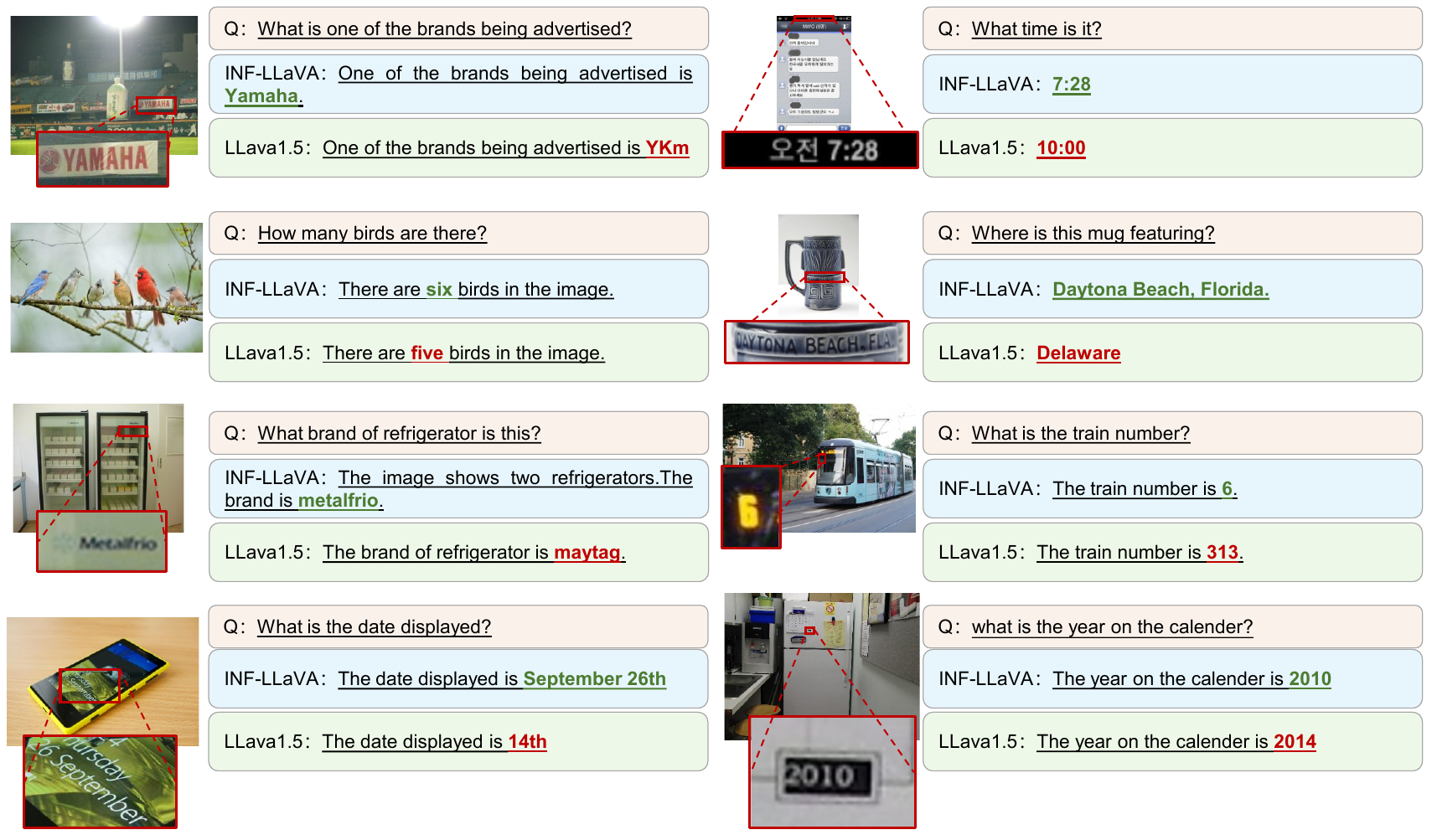}
\caption{Chat comparison between \modelname and LLaVA-1.5. Certain regions of the input high-resolution images are zoomed in for enhanced visualization.}
\label{fig:case2}
\end{figure*}

In Fig.~\ref{fig:case1} and Fig.~\ref{fig:case2}, we present a detailed comparison between \modelname and LLaVA-1.5, both of which utilize the same training dataset. The results clearly demonstrate that \modelname outperforms LLaVA-1.5 in several critical areas.
\textit{Firstly}, in terms of text recognition capabilities, \modelname exhibits greater accuracy. For instance, as shown in the first case of Fig.~\ref{fig:case1}, when asked, "What is the tagline with 'Wendell Rodricks' name?", \modelname accurately identifies the tagline, whereas LLaVA-1.5 provides an incorrect response. This highlights \modelname's superior ability to discern and interpret textual information within high-resolution images.
\textit{Secondly}, regarding the issue of hallucination, \modelname demonstrates a clear advantage. When uncertain about an answer, \modelname candidly indicates its uncertainty, whereas LLaVA-1.5 tends to fabricate responses, leading to hallucination problems. For example, in the last case of Fig.~\ref{fig:case1}, when asked to provide Chinese text, LLaVA-1.5 incorrectly responds with English text, whereas \modelname acknowledges its inability to answer, thereby avoiding erroneous information.
\textit{Thirdly}, in terms of counting accuracy, \modelname benefits from the enhanced details provided by high-resolution images, enabling more precise perception and reasoning. For instance, as depicted in the first case of Fig.~\ref{fig:case2}, \modelname correctly identifies the number of birds in the image, while LLaVA-1.5 fails to provide an accurate count. This demonstrates \modelname's superior capabilities in tasks requiring detailed visual analysis.
These comparisons consistently highlight the superior performance of \modelname across various complex tasks, emphasizing its advanced capabilities in text recognition, reducing hallucination issues, and enhancing counting accuracy.

\subsection{Ablation Studies}

\subsubsection{Diverse Image Resolutions}

\begin{figure*}[]
\centering
\includegraphics[width=2.0\columnwidth]{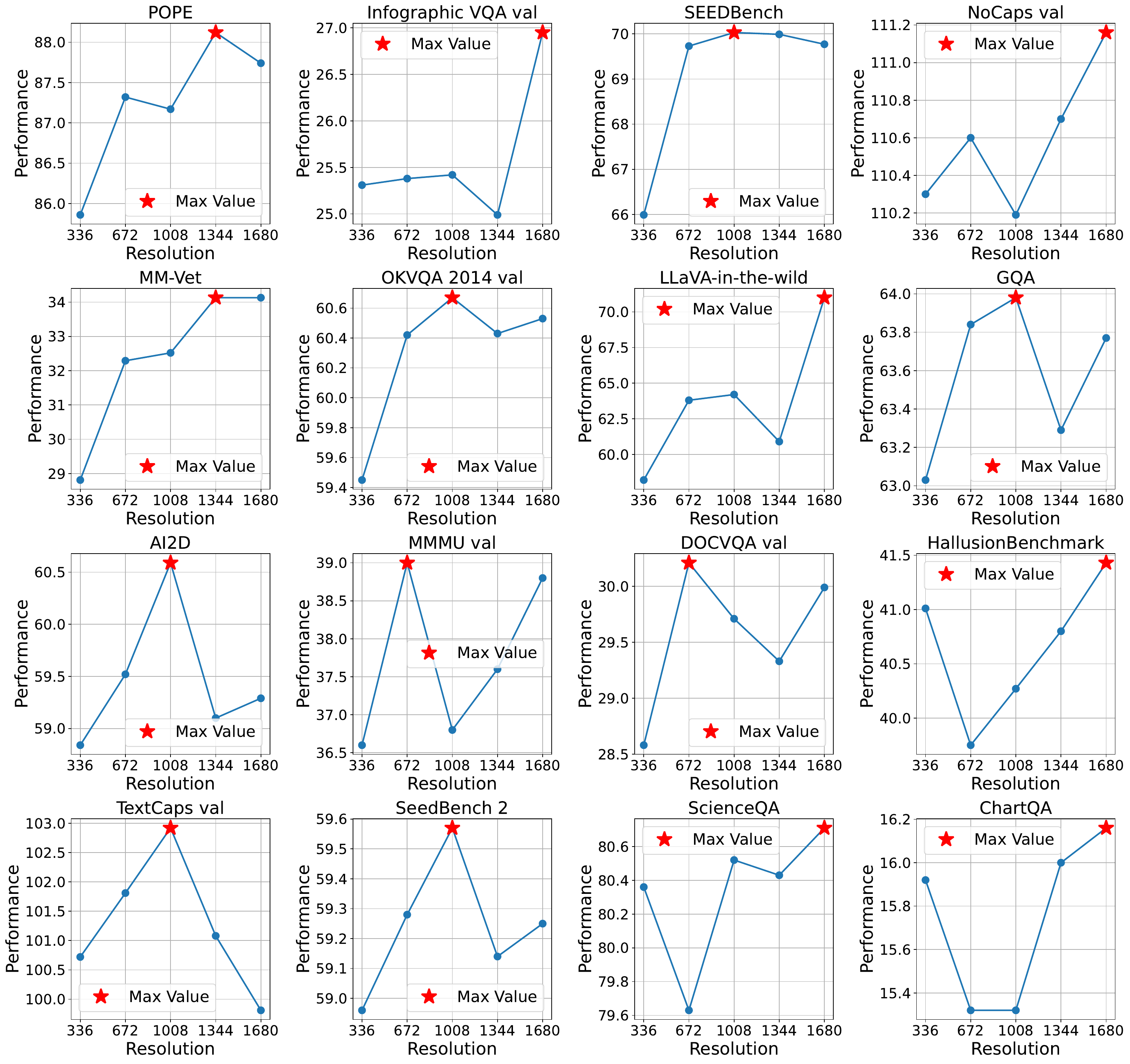}
\caption{Performance evaluation of \modelname across various benchmarks and image resolutions. The best performance for each benchmark is highlighted with a red star, illustrating the optimal resolution for \modelname in different contexts. Unlike the multi-resolution approach shown in the latter half of Tab.~\ref{tab:resolution}, this evaluation used a consistent single resolution for each experiment to explore the impact of resolution.}
\label{fig:resolution}
\end{figure*}

\begin{table*}[]
\caption{Performance on the POPE, SEEDBench, MM-Vet, LLaVA-Bench-in-the-wild benchmarks using different resolutions. }
\centering
\renewcommand\arraystretch{1.2} 
\begin{tabular}{l|cccc}
\toprule
\textbf{Resolution}     & \textbf{POPE}  & \textbf{SEEDBench} & \textbf{MM-Vet} & \textbf{LLaVA-wild} \\
\midrule
\textbf{{[}336{]}}      & 85.86          & 65.99              & 28.81           & 58.20               \\
\hline
\textbf{{[}672{]}}      & 87.32          & 69.73              & 32.29           & 63.80               \\
\textbf{{[}1008{]}}     & 87.17          & 70.03              & 32.52           & 64.20               \\
\textbf{{[}1344{]}}     & \textbf{88.12} & 69.99              & 34.13           & 60.90               \\
\textbf{{[}1680{]}}     & 87.74          & 69.77              & 34.13           & \textbf{71.00}      \\
\hline
\textbf{{[}336+672{]}}  & 87.61          & 69.94              & 31.74           & 59.40               \\
\textbf{{[}336+1008{]}} & 86.94          & \textbf{70.47}     & \textbf{34.50}  & 67.50               \\
\textbf{{[}336+1344{]}} & 87.36          & 69.04              & 34.04           & 61.50               \\
\textbf{{[}336+1680{]}} & 87.31          & 69.35              & 33.99           & 64.50     \\
\bottomrule
\end{tabular}
\label{tab:resolution}
\end{table*}

We conduct a series of experiments to evaluate the impact of different input image resolutions on the performance of \modelname, as detailed in Tab.~\ref{tab:resolution}. Initially, we evaluate the model using single-resolution inputs and observe that higher-resolution images generally yield better results compared to lower-resolution images (\emph{e.g.,} $336 \times 336$). For instance, on the POPE benchmark, \modelname achieves scores of 85.86, 87.32, 87.17, and 88.12 for resolutions of $336 \times 336$, $672 \times 672$, $1008 \times 1008$, and $1344 \times 1344$, respectively. This trend underscores the benefit of high-resolution images in capturing finer details, which enhances the model's performance.
However, it is noteworthy that the performance improvement is not strictly linear with increasing resolution. The optimal resolution varies across different benchmarks, potentially due to variations in image content and the inherent difficulty levels of the benchmarks. 
Inspired by previous works~\cite{li2024monkey,dong2024internlm}, we explore the synergy of integrating both low-resolution and high-resolution images to capture comprehensive global and local information. When combining low-resolution and high-resolution features, we perform an element-wise addition of the two feature sets. As demonstrated in Tab.~\ref{tab:resolution}, this dual-resolution approach enables \modelname to achieve its best performance on benchmarks such as SEEDBench and MMVet when using $336 \times 336$ and $1008 \times 1008$ resolutions concurrently.
The superior results obtained with this dual-resolution strategy highlight its effectiveness in balancing detailed visualization with global contextual understanding. Consequently, we adopt this resolution setting as the default configuration in subsequent ablation experiments. 

To further investigate the impact of varying image resolutions, we conducted a comprehensive set of experiments testing \modelname across multiple benchmarks with different image resolutions. As illustrated in Fig.~\ref{fig:resolution}, we observe a general trend where performance improves with increased resolution on most benchmarks, such as POPE~\cite{li2023evaluating}, Infographic VQA~\cite{mathew2022infographicvqa}, SEEDBench~\cite{li2023seed}, MM-Vet~\cite{yu2023mm}, OKVQA~\cite{marino2019ok}, and LLaVA-in-the-wild~\cite{liu2024visual}. This trend indicates that higher resolutions enable \modelname to capture more detailed visual information, thereby enhancing performance.
However, it is important to note that this upward trend does not uniformly apply across all benchmarks. For instance, on GQA~\cite{hudson2019gqa}, AI2D~\cite{kembhavi2016diagram}, and TextCaps~\cite{sidorov2020textcaps}, the optimal resolution is $1008 \times 1008$. On the MMMU~\cite{yue2024mmmu} and DOCVQA~\cite{mathew2021docvqa} benchmarks, the best performance is achieved at a resolution of $672 \times 672$. This variability suggests that increasing resolution does not necessarily lead to continuous performance improvement for all tasks. In certain cases, higher resolutions may introduce redundancies or distortions due to image enlargement, which can negatively affect model performance.


\subsubsection{Variants of \moduleTwoBig}

\begin{table*}[]
\caption{Performance comparison of different variants of \moduleTwoBig (\moduleTwoBigShort). \textbf{DEM w/ global} represents the module utilizing only Global-Perspective Enhancement, while \textbf{DEM w/ local} represents the module utilizing only Local-Perspective Enhancement. This comparison elucidates the individual impact of global and local enhancements on model performance.}
\centering
\renewcommand\arraystretch{1.2} 
\begin{tabular}{l|cccc}
\toprule
\textbf{}              & \textbf{POPE}  & \textbf{SEEDBench} & \textbf{MM-Vet} & \textbf{LLaVA-wild} \\
\midrule
\textbf{DEM w/ global} & 86.15          & 69.06              & 31.43           & 62.90               \\
\textbf{DEM w/ local}  & 86.24          & 69.45              & 31.66           & 62.70               \\
\textbf{DEM}           & \textbf{86.94} & \textbf{70.47}     & \textbf{34.50}  & \textbf{67.50}     \\
\bottomrule
\end{tabular}
\label{tab:dem}
\end{table*}

As illustrated in Fig.~\ref{fig:dem}, \moduleTwoBigShort integrates two crucial enhancement operations: Global-Perspective Enhancement and Local-Perspective Enhancement. In this section, we conduct a series of ablation experiments to rigorously evaluate the effectiveness of these two enhancements.

First, we evaluate \moduleTwoBigShort configured with only the global-perspective enhancement (\emph{i.e.,} DEM w/ global). The results, presented in the first line of Tab.~\ref{tab:dem}, show a significant reduction in performance across all benchmarks when the local-perspective enhancement is excluded. This clear decline in performance underscores the critical role that local-perspective enhancement plays in capturing fine-grained details necessary for high model accuracy.

Next, we test \moduleTwoBigShort with only the local-perspective enhancement (\emph{i.e.,} DEM w/ local). As depicted in the second line of Tab.~\ref{tab:dem}, DEM w/ local yields slightly better performance than DEM w/ global on the POPE, SEEDBench, and MM-Vet benchmarks. Conversely, it performs slightly worse on the LLaVa-Bench-in-the-wild benchmark compared to the DEM w/ global. However, it is important to highlight that both configurations fall short of the complete DEM, which includes both enhancement operations.

These ablation studies conclusively demonstrate that the combined use of both perspectives in \moduleTwoBigShort results in superior performance by effectively capturing unique and complementary aspects of the visual data. This integrative approach ensures a balanced and comprehensive understanding, ultimately leading to significantly enhanced model performance across a wide range of benchmarks.

\subsubsection{Fusion methods in \moduleTwoBigShort}

\begin{table*}[]
\caption{Comparison of different feature fusion methods in \moduleTwoBigShort. The performance is evaluated using various benchmarks. \textbf{Multiplication} denotes element-wise multiplication.}
\renewcommand\arraystretch{1.2} 
\centering
\begin{tabular}{l|cccc}
\toprule
\textbf{}                                                                       & \textbf{POPE}  & \textbf{SEEDBench} & \textbf{MM-Vet} & \textbf{LLaVA-wild} \\ 
\midrule
\textbf{Conv(3x3)}                                                              & 86.64          & 69.63              & 25.60           & 65.70               \\
\textbf{Multiplication} & 86.88          & 69.80              & 30.69           & 62.80               \\
\textbf{Weighted Addition}           & 86.21          & 69.61              & \textbf{33.53}  & 62.70               \\
\textbf{Maxpool}                                                                & 85.92          & 69.66              & 24.45           & 63.20               \\
\textbf{Addition}                                                               & 86.01          & 70.03              & 32.02           & 60.70               \\
\textbf{Linear-Concat}                                                                 & \textbf{87.04} & \textbf{70.62}     & 33.44           & \textbf{66.40}      \\ 
\bottomrule
\end{tabular}
\label{tab:fusion}
\end{table*}

Fig.~\ref{fig:dem} illustrates the process of obtaining dual-enhanced features, which necessitate an effective fusion method to combine the local-enhanced feature and global-enhanced feature. In our proposed method, we utilize linear layers to reduce the dimensionality of the features, followed by concatenation to form the dual-enhanced features. 
This section explores the impact of various fusion methods on performance. Different techniques, including $3 \times 3$ convolution, element-wise multiplication, weighted addition, max pooling, and addition, were employed to fuse the features. As shown in Tab.~\ref{tab:fusion}, the experimental results provide a comprehensive comparison of these methods across several benchmarks. 
Notably, the embed-and-concat method (Linear-Concat) achieves superior performance on most benchmarks. It outperforms other methods, achieving the highest scores on the POPE, SEEDBench, and LLaVA-wild benchmarks. This indicates that by carefully embedding and concatenating the features, our method effectively integrates the global and local perspectives.


\subsubsection{Module Ablation}

\begin{table*}[]
\caption{Ablation studies of the proposed \moduleTwoBigShort module. \textbf{w/ DCM local} indicates the use of \moduleTwoBigShort with only local-perspective cropping, omitting global-perspective cropping. \textbf{w/ DCM global} indicates the use of \moduleTwoBigShort with only global-perspective cropping, omitting local-perspective cropping. \textbf{w/ DCM} indicates the approach where local-perspective and global-perspective features are combined through element-wise addition.}
\centering
\renewcommand\arraystretch{1.2} 
\begin{tabular}{l|cccc}
\toprule
\textbf{}              & \textbf{POPE}  & \textbf{SEEDBench} & \textbf{MM-Vet} & \textbf{LLaVA-wild} \\
\midrule
\textbf{w/ DCM local}  & 86.12          & 66.60              & 27.80           & 56.30               \\
\textbf{w/ DCM global} & 86.15          & 70.06              & 32.43           & 62.90               \\
\textbf{w/ DCM}        & 85.77          & 68.89              & 18.90           & 60.10               \\
\textbf{w/ DCM+DEM}    & \textbf{87.04} & \textbf{70.62}     & \textbf{33.44}  & \textbf{66.40}      \\
\bottomrule
\end{tabular}
\label{tab:module}
\end{table*}

In this section, we conduct a thorough ablation study to examine the effectiveness of the proposed modules. As detailed in Tab.~\ref{tab:module}, we first evaluate the performance when using only one type of cropping method. The first two lines show that using global-perspective cropping alone (\emph{w/ DCM global}) consistently outperforms local-perspective cropping alone (\emph{w/ DCM local}). This result suggests that global-perspective cropping, which maintains continuous visual information, is beneficial for model performance. In contrast, local-perspective cropping may inadvertently segment complete objects into several sub-images, complicating the recognition and understanding processes.
Next, we investigate the impact of combining both cropping features through element-wise addition of the features (\emph{w/ DCM}) instead of \moduleTwoBigShort. As evident from the third line of Tab.~\ref{tab:module}, this approach results in even lower performance than using only one type of cropping. We hypothesize that this degradation in performance arises due to the differing information densities of the tokens produced by each cropping method, making direct element-wise addition a suboptimal fusion strategy.
Finally, we consider the complete \moduleTwoBigShort module, which integrates both cropping methods in a more sophisticated manner. The last row of Tab.~\ref{tab:module} demonstrates that \moduleTwoBigShort achieves the highest performance across all benchmarks. This clearly indicates that the combined use of local- and global-perspective enhancements, when managed effectively, significantly improves model performance. These findings underscore the effectiveness of the proposed \moduleOneBigShort and \moduleTwoBigShort modules and highlight the importance of thoughtful feature integration strategies in enhancing model capabilities.

\section{Conclusion}

In this paper, we proposed \modelname, a novel MLLM designed for high-resolution image perception and reasoning. \modelname leverages two innovative modules: \moduleOneBig (\moduleOneBigShort), which crops high-resolution images into sub-images from both local and global perspectives, and \moduleTwoBig (\moduleTwoBigShort), which efficiently fuses these features to obtain dual-enhanced features. Extensive experiments demonstrate that these modules significantly enhance \modelname's ability to understand high-resolution images, resulting in outstanding performance across various benchmarks. This work establishes a new state-of-the-art in vision-language tasks.

{\small
\bibliographystyle{ieee}
\bibliography{egbib}
}

\end{document}